\documentclass[
]{ceurart}

\usepackage{listings}
\begin{document}

\copyrightyear{2025}
\copyrightclause{Copyright for this paper by its authors.
  Use permitted under Creative Commons License Attribution 4.0
  International (CC BY 4.0).}

\conference{CLEF 2025 Working Notes, 9 -- 12 September 2025, Madrid, Spain}


\title{Zero-Shot Segmentation through Prototype-Guidance for Multi-Label Plant Species Identification}

\author[1]{Luciano Araujo Dourado Filho}[%
orcid=0000-0002-0507-2201,
email=lucianoadfilho@gmail.com,
url=https://github.com/FalsoMoralista,
]
\cormark[1]
\address[1]{ADAM - Advanced Data Analysis and Management, University of Feira de Santana (UEFS), Feira de Santana, Brazil}

\author[1]{Almir Moreira da Silva Neto}[%
orcid=0009-0008-5042-5556,
email=almirneto338@gmail.com,
url=https://github.com/almirneeto99/,
]

\author[2]{Rodrigo Pereira David}[%
orcid=0000-0001-9218-8191,
email=rpdavid@inmetro.gov.br,
url=https://github.com/rpdavid78/,
]
\address[2]{National Institute of Metrology, Technology and Quality (Inmetro)}

\author[1]{Rodrigo Tripodi Calumby}[%
orcid=0000-0001-8515-265X,
email=rtcalumby@uefs.br,
url=https://www.rtcalumby.com.br]

\cortext[1]{Corresponding author.}

\begin{abstract}
    This paper presents an approach developed to address the PlantClef 2025 challenge, which consists of a fine-grained multi-label species identification, over high-resolution images. Our solution focused on employing class prototypes obtained from the training dataset as a proxy guidance for training a segmentation Vision Transformer (ViT) on the test set images. To obtain these representations, the proposed method extracts features from training dataset images and create clusters, by applying K-Means, with $K$ equals to the number of classes in the dataset. The segmentation model is a customized narrow ViT, built by replacing the patch embedding layer with a frozen DinoV2, pre-trained on the training dataset for individual species classification. This model is trained to reconstruct the class prototypes of the training dataset from the test dataset images. We then use this model to obtain attention scores that enable to identify and localize areas of interest and consequently guide the classification process. The proposed approach enabled a domain-adaptation from multi-class identification with individual species, into multi-label classification from high-resolution vegetation plots. Our method achieved fifth place in the PlantCLEF 2025 challenge on the private leaderboard, with an F1 score of 0.33331. Besides that, in absolute terms our method scored 0.03 lower than the top-performing submission, suggesting that it may achieved competitive performance in the benchmark task. Our code is available at \href{https://github.com/ADAM-UEFS/PlantCLEF2025}{https://github.com/ADAM-UEFS/PlantCLEF2025}.
\end{abstract}


\maketitle

\section{Introduction}
The automated identification of plant species from images of complex vegetation plots presents a significant challenge in ecological research and biodiversity monitoring. Initiatives like the PlantCLEF aim to advance this field by providing benchmark datasets and tasks, fostering innovation in multi-label classification from high-resolution plot imagery~ \cite{Goeau2024plantclef_overview, Joly2024lifeclef_overview, lifeclef2025, plantclef2025}. A primary challenge in this task is the notable domain shift between training data, which mostly consist of images of individual plants, and test data, comprising dense, multi-species vegetation plots captured under varied environmental conditions~ \cite{lifeclef2025,plantclef2025,Chulif2024neuon, Foy2024atlantic}. Therefore, it demands robust methods that can effectively generalize and differentiate target species within cluttered natural scenes~ \cite{Bao2023transfer}.

Vision Transformers (ViTs), particularly those pre-trained using self-supervised learning, have demonstrated remarkable effectiveness in generating discriminative visual features for various tasks, including large-scale plant identification~ \cite{Caron2021emerging, Kattenborn2021cnn}. In the broad context of image classification, models such as DINOv2~\cite{Oquab2023dinov2} has demonstrated to offer powerful backbones for feature extraction. Specifically in The PlantCLEF 2025, organizers supported participants by providing ViT models pre-trained on relevant flora, considering a DINOv2-based architectures~ \cite{Goeau2024models}. Our proposed method goes beyond the usual direct end-to-end classification of the entire plot images (which can be confounded by background noise and species overlap) by focusing on a preliminary segmentation-driven filtering step. We propose a ``decrease-and-conquer'' strategy which relies in a dedicated ViT that first identifies plant-relevant regions, thereby simplifying the subsequent classification task for specialized heuristics. This ViT leverages its attention mechanisms as a proxy for spatial segmentation, aiming to isolate pertinent areas from irrelevant background elements.

A crucial aspect of our methodology is the training of this segmentation ViT without direct pixel-level supervision. To overcome the absence of segmentation labels, we introduce a novel proxy task as our core contribution. We trained a narrow, customized ViT to reconstruct a predetermined representation of the \emph{training dataset}---specifically, class prototypes derived from k-Means clustering on DINOv2 embeddings of the original training set images. Uniquely, this ViT learns this reconstruction objective using, as its input, feature embeddings extracted by a separate, frozen DINOv2 model from \emph{test images}. Our hypothesis is that for the ViT to successfully map features from unseen test plots back to these known training set prototypes, it must learn to generate attention scores that highlight plant-relevant patches while attenuating signals from the background (as illustrated in Figure~\ref{fig:quality_attention_maps}).

This paper provides a detailed exposition of this proxy task-driven ViT technique. The generation of the target class prototypes, the specific architecture of the custom ViT, and the operational steps of its unique training regimen as detailed in Section~\ref{sec:method}. Furthermore, we describe how the learned attention maps are utilized within our inference pipeline to select regions of interest prior to applying classification heuristics for final species identification. The experimental settings, key hyperparameters, and the results achieved in the PlantCLEF 2025 challenge are presented and discussed, offering insights into the effectiveness of this approach for multi-label classification in complex ecological imagery.







\section{Task Description}
The PlantCLEF 2025 challenge addresses the intricate problem of multi-label species identification within high-resolution images of vegetation plots. The core objective is to predict all plant species present in a given plot image, a fundamental task to ecological research, biodiversity assessment, and long-term environmental monitoring.

The primary difficulty of the challenge stems from a significant domain shift between the provided datasets. A comprehensive training dataset comprising roughly 1.4 million images representing 7,806 plant species. These training images typically feature single plant species from various perspectives. In contrast, the test dataset consists of 2,105 high-resolution plot images from Mediterranean and Pyrenean regions. These test images depict complex, multi-species scenes, often captured from a top-down perspective within delimited sampling areas (quadrats), and under varying environmental conditions that can introduce shadows or blurry areas.

Two pre-trained ViT models based on the DINOv2 architecture were provided along with the dataset. These models were not only pre-trained on a massive dataset of images but were also specifically fine-tuned on the PlantCLEF 2024 training data which includes isolated plant samples, considering multiple organs and perspectives. The fine-tuned models available were: a) one where only the final classification layer was trained; b) another where all model layers were unfrozen and fine-tuned for the task. These models offer a powerful, ready-to-use foundation resources avoiding new investigations to demand the extensive computational resources needed to train such large-scale models from scratch. The classification effectiveness was evaluated using the macro F1 score averaged per plot, a metric designed to balance precision and recall for each individual plot images, with results tracked on a public leaderboard. The final score is computed as described in Equation \ref{eq:final-score}, where $N$ is the number of transects (plots) , $T^{i}$ is the number of quadrats in transect $i$, $\frac{1}{T_{i}}\sum_{j=1}^{T_{i}}F1^{j}$ is the macro-averaged F1-score per sample of transect $i$ and $F1^{j}$ is the F1 score for test image $j$.

\begin{equation} \label{eq:final-score}
    Score = \frac{1}{N} \sum_{i=1}^{N}(\frac{1}{T_{i}}\sum_{j=1}^{T_{i}}F_1^{j})
\end{equation}

For each test image $j$ the F1 score is described as follows in Equation \ref{eq:f1-score}.

\begin{equation} \label{eq:f1-score}
    Macro \ F_1^{j} = \frac{2 \times Precision_{j} \times Recall_{j}}{Precision_{j} + Recall_{j}}
\end{equation}

The $Precision$ and $Recall$, for each test image $j$ is represented in Equation \ref{eq:precision} and \ref{eq:recall}, respectively.

\begin{equation} \label{eq:precision}
    Precision = \frac{TP_{j}}{TP_{j} + FP_{j}}
\end{equation}

\begin{equation} \label{eq:recall}
    Recall = \frac{TP_{j}}{TP_{j}+FN_{j}}
\end{equation}

Where $TP_{j}$ is the number of species correctly predicted, $FP_{j}$ is the number of plant species incorrectly predicted and $FN_{j}$ number of plant species missed.

\section{Proposed Method and Experiments}
\label{sec:method}
\begin{figure}[b]
    \centering
    \includegraphics[width=0.99\columnwidth]{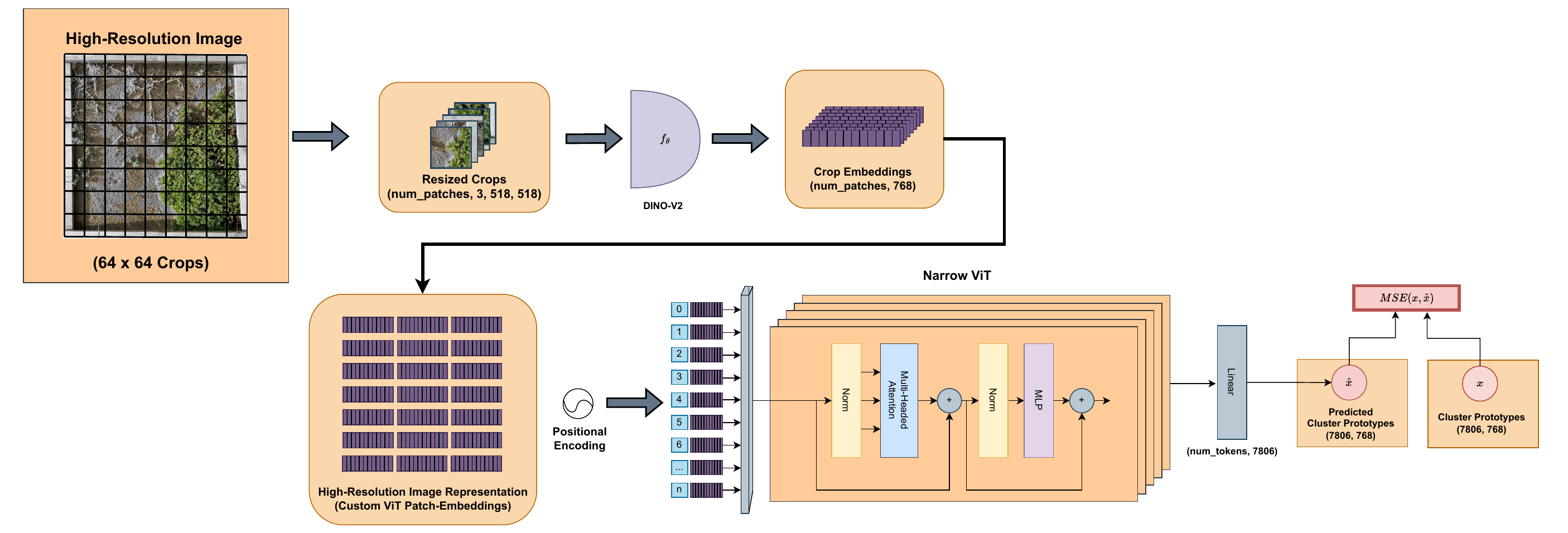}
    \caption{Training process diagram.}
     \label{fig:training_step}
\end{figure}
The proposed zero-shot learning approach focuses on the segmentation of plants from irrelevant background elements within test images. In order to achieve that, we trained a dedicated segmentation model, whose primary function is to identify plant regions. We theorize that a preliminary segmentation step could allow for more precise application of classification heuristics, over pertinent regions in the image, in a "decrease-and-conquer" fashion.

For the segmentation task, we employed a ViT due to its capability to model contextual relationships between image patches (tokens), according to an objective. We propose that given the appropriate objective, these relationships could be guided towards enabling the model to differentiate plant structures from other objects. In other words, we anticipate that the ViT's attention scores could provide a mechanism for selecting relevant regions from test images. 

The caveat of our approach is that in order for this model to learn semantically meaningful (patch-wise) relationships, a reasonable objective has to be given, which we initially do not have. To tackle such limitation, a proxy task was designed consisting of reconstructing an approximate representation of the training dataset. We used the training dataset to obtain features that represented each of its individual species generating cluster prototypes, which we employ as reconstruction objective for our ViT. We considered that in order to successfully reconstruct the training dataset representation from test-set images, the model would have to learn patch-wise relationships (attention scores) that enabled the maximization of this objective. As a consequence, our assumption is that the model would learn to assign low attention scores to irrelevant regions (background noise) and high scores to patches containing plants (as illustrated in Figure~\ref{fig:training_step}).

\subsection{Clustering as proxy task}

To obtain the target representation that guides the ViT's optimization, we applied K-Means clustering to the image embeddings of the training dataset. These embeddings were extracted using a separate, identical instance of the pretrained DINOv2 model, from which the classification head has been removed. Following embedding extraction, K-Means was performed with K set to the number of classes (7806 species). The resulting cluster centroids, or 'class prototypes', constitute the predetermined representation of the training dataset that the narrow ViT is addressed to reconstruct. Figure~\ref{fig:training_step} illustrates the complete training process wherein the ViT learns this reconstruction.

Figure~\ref{fig:training_step} details the operational steps for training our narrow ViT on this proxy reconstruction objective. The process begins with resizing each image from the test dataset into a predefined $H \times W$ resolution, where H and W denote image height and width. Following that, each high-resolution image is segmented into a grid of non-overlapping $64 \times 64$ pixel patches (crops). These patches are subsequently resized to $518 \times 518$ pixels using bicubic interpolation to meet the input specifications of the pre-trained DINOv2 model, which serves as the patch feature extractor. This procedure yields $num\_patches = \frac{H \times W}{64^2}$ tokens per image.

These resized $518 \times 518$ pixel crops from the test images are then processed by the aforementioned frozen DINOv2 model to extract the corresponding (768-dimensional) feature embeddings. Positional encodings are subsequently integrated with these patch embeddings for the ViT processing. This final sequence of augmented patch embeddings serves as the input to the narrow ViT. The ViT architecture is a ViT-Base (ViT-B) variant configured with 6 transformer blocks, 12 attention heads per block, an embedding dimension of 768, and a linear layer. This model is optimized to reconstruct the predetermined target matrix of class prototypes (dimensions $7806 \times 768$) from these input embeddings.

\subsection{Inference}\label{sub_inference}

Following the training procedure, an optimized Vision Transformer (ViT) is employed to generate attention maps for the test images. These scores facilitate the removal of background and other irrelevant elements from images, prior to the classification stage. To accomplish this, an image is first processed through the ViT. The attention scores from all transformer blocks are then aggregated by averaging them across both blocks and attention heads. This averaged result is subsequently normalized to create a final attention map. This attention map guides the selection of relevant image regions by filtering tokens: only tokens whose normalized attention scores exceed a predefined threshold $t$ are retained. After this filtering step, which discards potentially irrelevant elements, the process advances to the species identification pipeline. For this identification task, we implemented two distinct procedures which are detailed in the following. 

\begin{figure}[htbp]
    \centering
    \includegraphics[width=0.95\columnwidth]{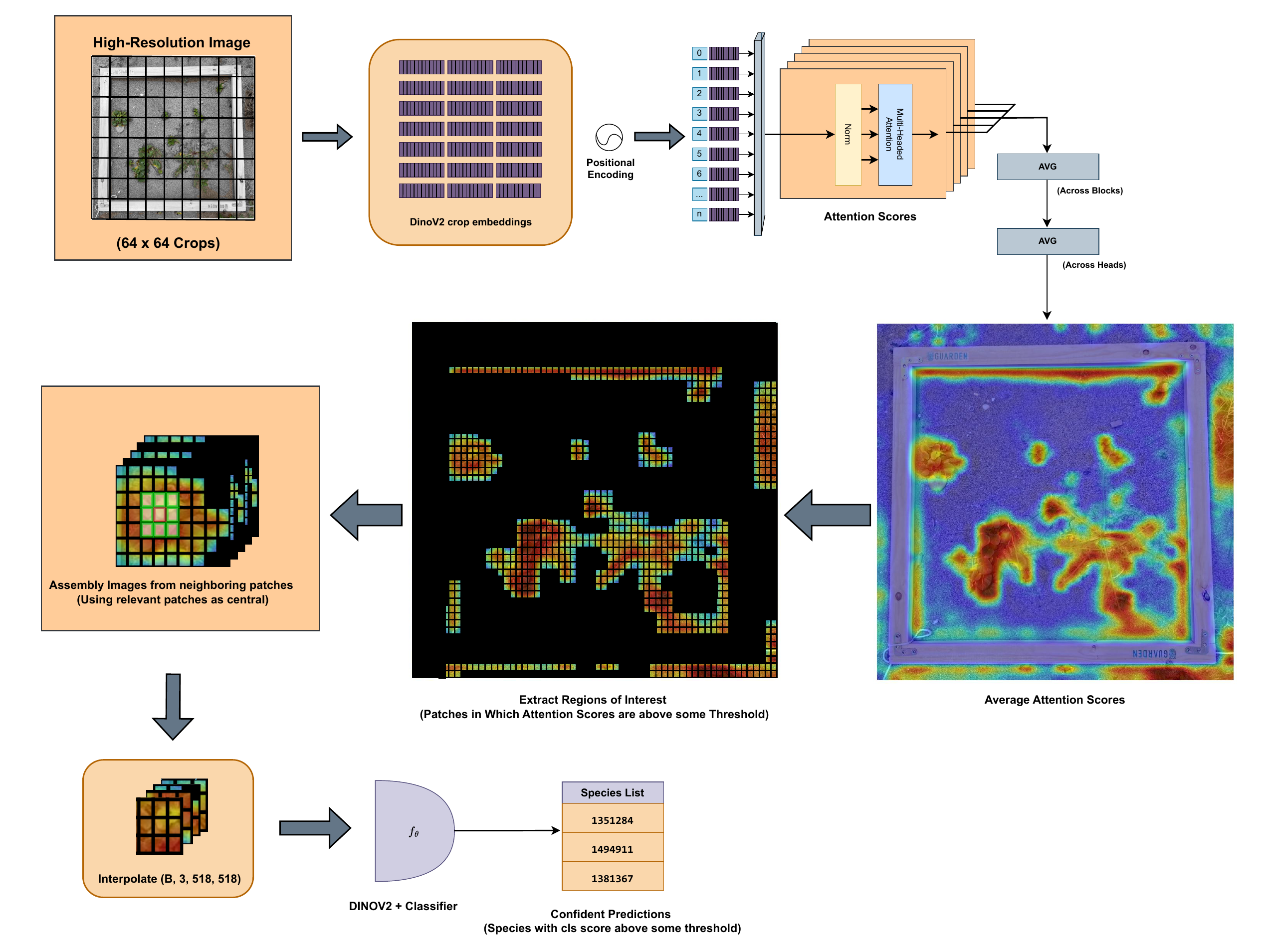} 
    \caption{Inference stage diagram.}
     \label{fig:inference_time}
\end{figure}


As an initial step, the first heuristic comprises resizing each relevant crop to conform to the input dimensions required by the pretrained DINOv2 model. Following this preprocessing, each resized crop serves as input to the pretrained DINOv2 (with classifier) to obtain the classification scores for each crop. From that, the predicted species was determined by selecting the most confident prediction that surpassed a predefined threshold: $prob$. In instances where no prediction met this criterion, the species associated with the highest overall probability score was selected by default. The final list of predicted species for a given quadrat was then compiled by aggregating the unique species IDs from all individual crop-level predictions within that quadrat.

Also following the application of a threshold ($t$) to the normalized attention map, the second heuristic consisted of constructing a composite image as follows. For each patch with an attention score above~$t$, an image was assembled as a~$K \times K$~grid of neighbouring patches, using the relevant patch as the central element. This newly assembled image was then resized via bicubic interpolation to match the input dimensions of the pretrained DINOv2 model and processed for classification. Analogous to the first heuristic, this approach assumes the classifier can inherently filter irrelevant or outlier regions by prioritizing predictions that achieve a confidence score above a specified threshold (\textit{prob}). Figure~\ref{fig:inference_time} illustrates this procedure, which was the approach that yielded the best performance of the proposed method. During this inference stage, we experimented over the classification parameters depicted in Table~\ref{tab:cls_parameters} to optimize predictive performance.

\subsection{Experimental Settings}
Our custom ViT implementation is based on the I-JEPA architecture~\cite{ijepa}, in which the \texttt{[cls]} token is disregarded. We evaluated two input image resolutions: $2048 \times 2048$ pixels and $3072 \times 2048$ pixels. When processed into a grid of non-overlapping $64 \times 64$ pixel crops, these resolutions correspond to 1024 and 1536 patches (tokens) per image, respectively.

\begin{table} [b]
    \caption{Classification parameters values}
    \centering
    \begin{tabular}{cc}
        \hline
         Classification Parameter &  Range \\
         \hline
         Attention threshold (t) & [0.5, 0.6, 0.7] \\
         \hline
         Probability threshold (prob) & [0.3, 0.45 ,0.5, 0.6, 0.7, 0.8] \\
         \hline
         Grid Size (K) & [5, 9] \\ 
         \hline
    \end{tabular}
    \label{tab:cls_parameters}
\end{table}

Model optimization was performed for up to 30 epochs using the AdamW optimizer. This included a cosine learning rate decay schedule~\cite{ijepa}, typically applied over 10 to 20 epochs of the training period, and a weight decay schedule that linearly increased from $0.04$ to $0.4$. A consistent learning rate profile was adopted across all experiments, defined by: a starting learning rate ($LR_{\text{start}}$) of $5.0 \times 10^{-6}$, an effective peak learning rate ($LR_{\text{effective}}$) of $1.0 \times 10^{-3}$, and a final learning rate ($LR_{\text{final}}$) of $1.0 \times 10^{-6}$. Early stopping was implemented, triggered by observed degradation in the quality of attention maps and the onset of model overfitting. Ablation studies, extended to 100 epochs, confirmed that training for 30 epochs with this early stopping criterion was sufficient to achieve optimal performance.

All experiments were conducted on a system equipped with two NVIDIA A100-SXM 80GB GPUs. A per-GPU batch size of 128 was employed, achieved through gradient accumulation. Our observations indicated that this VRAM capacity constrained the maximum number of tokens processed simultaneously to approximately 2048 for the given model configuration.

\section{Results}
Table~\ref{tab:result_all_runs} summarizes the results all the configurations evaluated. Configurations 1-4 used our first classification strategy, which consisted of performing direct patch-wise classification over resized crops.

As demonstrated in Table~\ref{tab:result_all_runs}, the patch-wise classification strategy yielded the lowest performance among all evaluated. Specifically, after applying an attention threshold of~$t=0.6$ and classifying the resulting patches without any probability constraints (Configuration 1), the model achieved an $F_1$-score of 0.02114 on the private set. While a marginal improvement to an $F_1$-score of 0.09511 was obtained by introducing a probability threshold of $prob=0.5$ (Submission 3), the overall strategy demonstrated a low performance. We attribute these suboptimal results to the low-resolution nature of the individual patches, which likely misses the necessary contextual information required for the model to identify a complete plant specimen in a effective way.

In contrast, the alternative approach, adopting the strategy of assembling a contextual grid of adjacent patches around each relevant token (as depicted in Figure~\ref{fig:inference_time}) yielded a remarkable performance improvement. Using the exact same model weights, this strategy increased the $F_1$-score from 0.09511 to 0.33131. This outcome corroborates our hypothesis that individual, low-resolution patches lack sufficient contextual information for effective classification. Subsequent experiments with a smaller grid size ($K=5$ in submissions 15 and 16) provided further evidence, demonstrating that reducing the amount of contextual information by using fewer adjacent patches led to deterioration in the classification performance. More information regarding checkpoint epoch selection is described in Section~\ref{subsec:training_dynamics}

\begin{table}[htbp]
    \centering
    \caption{Performance metrics of the evaluated 16 configurations with varying hyperparameter configurations. The configurations are grouped by classification strategy. The absence of parameter $K$ indicates the use of the direct patch-wise classification strategy (Heuristic 1). Public Score results are reported on 11\% of the data, whereas the Private Score is obtained from the remaining 89\%.}
    \label{tab:result_all_runs}
    \begin{tabular}{cccccccc}
        \hline
        Configuration & Input Resolution & Checkpoint Epoch & t   & prob & K & Public Score & Private Score  \\
        \hline
        \multicolumn{8}{c}{\textit{Patch-wise Classification (Heuristic 1)}} \\
        \hline
        1  & $2048 \times 2048$ & 15 & 0.6 & -    & - & 0.02709 & 0.02114 \\
        \hline
        2  & $2048 \times 2048$ & 15 & 0.6 & 0.15 & - & 0.08554 & 0.07328 \\
        \hline
        3  & $2048 \times 2048$ & 15 & 0.6 & 0.5  & - & 0.11809 & 0.09511 \\
        \hline
        4  & $2048 \times 2048$ & 15 & 0.6 & 0.7  & - & 0.13248 & 0.09367 \\
        \hline
        \multicolumn{8}{c}{\textit{Grid Assembly Classification (Heuristic 2)}} \\
        \hline
        5  & $2048 \times 2048$ & 15 & 0.6 & 0.5  & 9 & 0.32606 & 0.33131 \\
        \hline
        6  & $2048 \times 2048$ & 15 & 0.6 & 0.8  & 9 & 0.26701 & 0.32461 \\
        \hline
        7  & $3072 \times 2048$ & 10 & 0.6 & 0.6  & 9 & 0.33069 & 0.33007 \\
        \hline
        8  & $3072 \times 2048$ & 10 & 0.6 & 0.7  & 9 & 0.31004 & 0.32181 \\
        \hline
        9  & $3072 \times 2048$ & 10 & 0.6 & 0.5  & 9 & \textbf{0.34427} & 0.33183 \\
        \hline
        10 & $3072 \times 2048$ & 10 & 0.6 & 0.45 & 9 & \textbf{0.34427} & 0.33183 \\
        \hline
        11 & $3072 \times 2048$ & 10 & 0.6 & 0.3  & 9 & \textbf{0.34427} & 0.33183 \\
        \hline
        12 & $3072 \times 2048$ & 10 & 0.5 & 0.5  & 9 & 0.34060 & 0.33151 \\
        \hline
        13 & $3072 \times 2048$ & 10 & 0.5 & 0.7  & 9 & 0.31153 & 0.32595 \\
        \hline
        \textbf{14} & \textbf{$3072 \times 2048$} & \textbf{10} & \textbf{0.7} & \textbf{0.5} & \textbf{9}  & 0.33867 & \textbf{0.33331} \\
        \hline
        15 & $3072 \times 2048$ & 10 & 0.6 & 0.60 & 5 & 0.30671 & 0.28580 \\
        \hline
        16 & $3072 \times 2048$ & 10 & 0.6 & 0.8  & 5 & 0.29442 & 0.28908 \\
        \hline
    \end{tabular}
\end{table}

The best results were obtained with configuration 14, where the segmentation model was trained with an input resolution of $3072 \times 2048$. This submission achieved F1$=0.33331$ on the private score, which represented an improvement of approximately 0.5\% over our first submission with this strategy (with an input resolution of $2048 \times 2048$ pixels). This findings indicated that increasing the input resolution may not be worthy, specially considering the computational and memory overhead associated to token processing in the ViT.

In summary this decrease-and-conquer strategy enabled the proposed method to reach the 5th place on the private leaderboard of PlantCLEF 2025. Moreover, our best configuration achieved a $F_1$-score of 0.33331, in contrast to the best performing proposal in the leaderboard, which achieved 0.36479. In absolute terms, our strategy presented roughly  a 0.03 difference in contrast to the first place, which demonstrates its high competitiveness. We believe that with further refinements, this strategy holds the potential to achieve even higher performance in future iterations. 

\subsection{Training Dynamics and Model Selection}
\label{subsec:training_dynamics}

\begin{figure}[htbp]
    \centering
    \includegraphics[width=0.95\columnwidth]{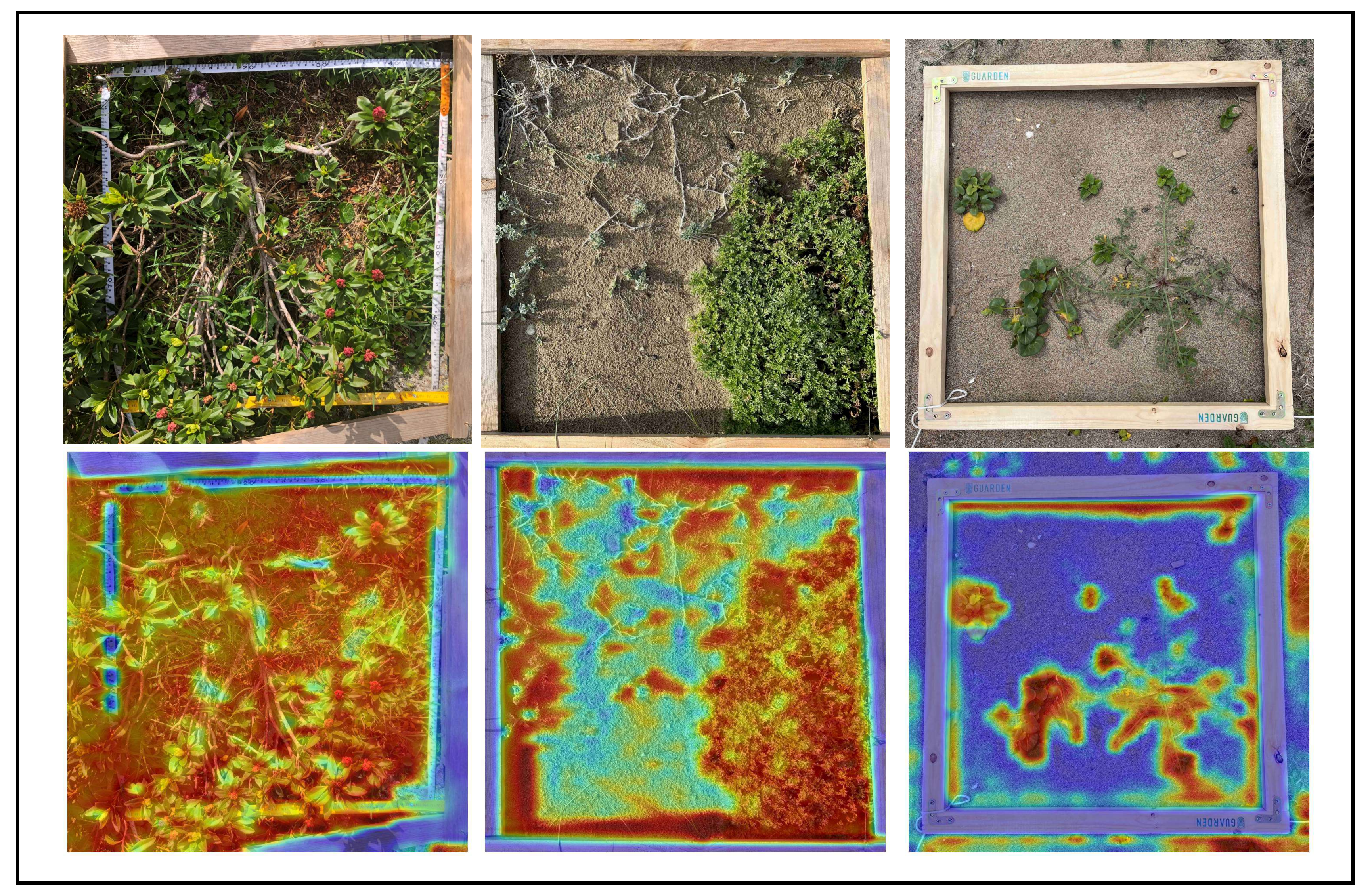}
    \caption{Qualitative results from the segmentation ViT during an early training epoch. The model correctly assigns high attention scores (red) to plant-containing regions and low scores (blue) to background elements, including the quadrat frame.}
    \label{fig:quality_attention_maps}
\end{figure}

Figure~\ref{fig:quality_attention_maps} shows a qualitative example of a well-formed attention map from an early training stage, where the model correctly learns to assign low attention scores (in blue) to irrelevant regions, such as the quadrat frame, and higher scores (in red) to regions containing plants.

\begin{figure}[htbp]
    \centering
    \includegraphics[width=0.95\linewidth]{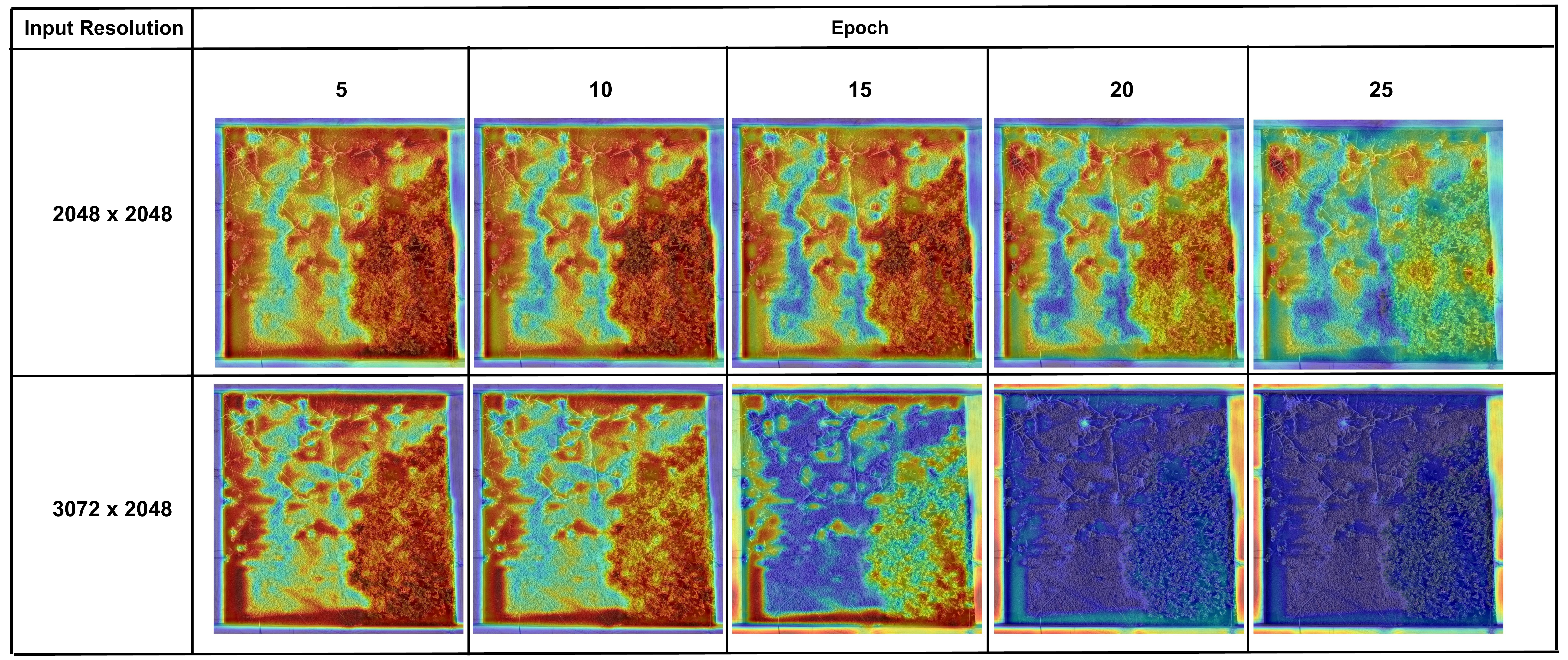}
    \caption{Illustration of attention map degradation over several training epochs. While initial epochs exhibit focused attention on plant regions, later epochs show a diffusion of attention and an incorrect intensification on background areas.}
    \label{fig:resulting_attn_maps}
\end{figure}

Regarding to model training, a counter-intuitive phenomenon was observed: a clear degradation in the quality of the attention maps coincided with the convergence the loss function (Figure~\ref{fig:resulting_attn_maps}).

\begin{figure}[htbp]
    \centering
    \includegraphics[width=0.75\linewidth]{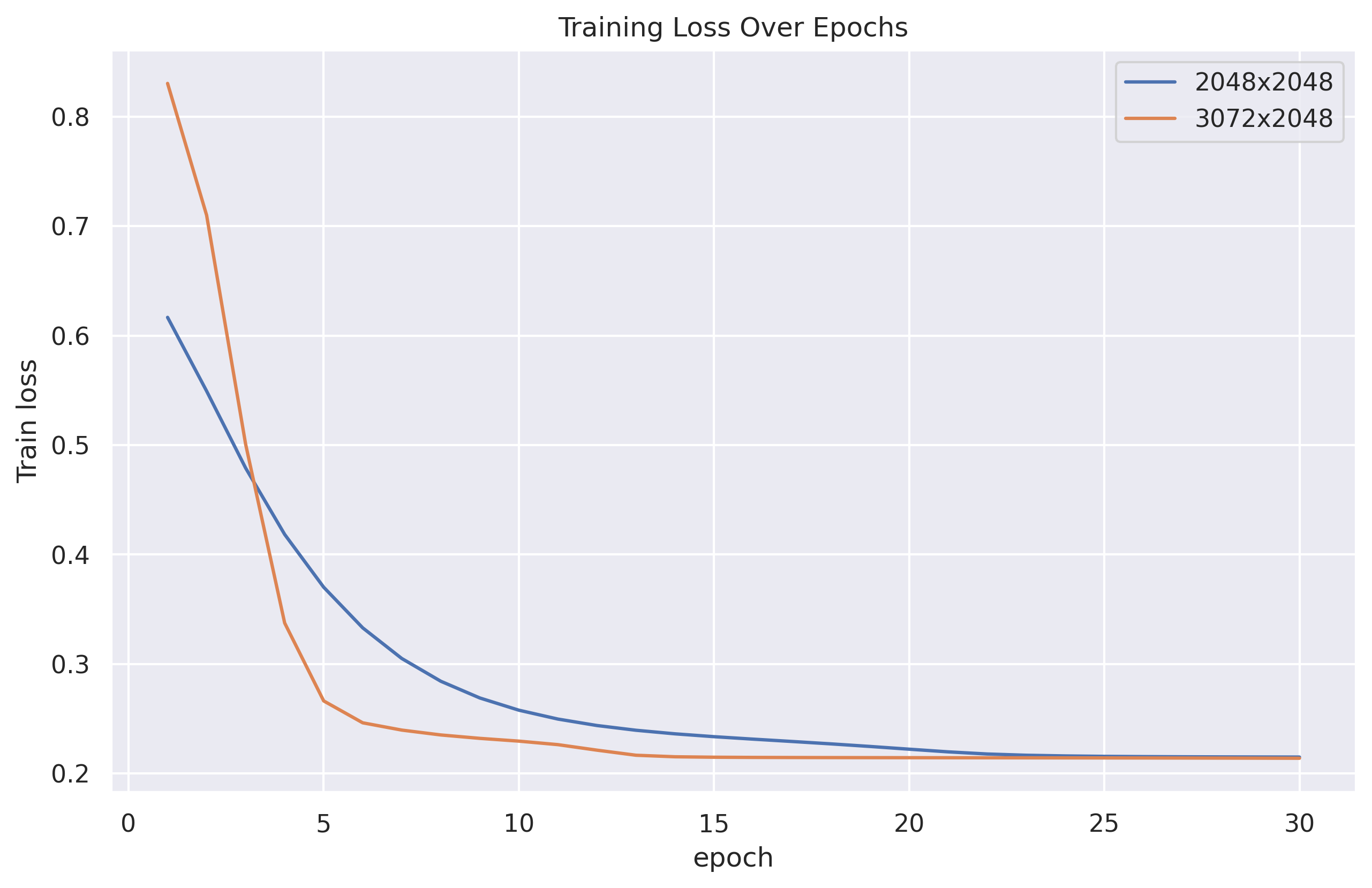}
    \caption{Training loss curve over epochs. The onset of loss convergence (plateau) aligns with the observed degradation in attention map quality, suggesting the model began to overfit to the reconstruction task.}
    \label{fig:training_losses}
\end{figure}

For the $2048 \times 2048$ input resolution, for example, the model performed as intended, up to approximately Epoch 15. Beyond this point, a clear reversal in behavior was noted: attention scores for relevant plant areas began to decrease, while scores for background elements started to intensify. An analysis of the training loss curve (Figure~\ref{fig:training_losses}) revealed that this deterioration coincided precisely with the onset of loss convergence. The same behavior was observed for the $3072 \times 2048$ resolution, although the degradation began earlier, after Epoch 10. This analysis suggested that later checkpoints could represent overfitted models with less meaningful semantic attention relationships. Because of that, we selected the checkpoints from Epoch 15 (for the $2048 \times 2048$ model) and Epoch 10 (for the $3072 \times 2048$ model) for all inference tasks presented in Table~\ref{tab:result_all_runs}.

In other words, we observed that although the model initially learns correct semantical relations, focusing its attention on plant features to reconstruct the class prototypes, as training progresses, it collapses by focusing on semantically irrelevant features, such as the quadrat frame. Our hypothesis is that this model collapse is somewhat expected, specially considering the objective of reconstructing a constant target matrix.

A possible conclusion is due a divergence between the optimization of an explicit reconstruction objective and the implicit goal of semantic segmentation, which without the appropriate regularization leads to the trivial solution of focusing on semantically irrelevant features. As a matter of fact, we anticipated that the objective of reconstructing a constant target matrix could lead to the straightforward trivial solution of the model ignoring the inputs and producing the outputs that perfectly reconstructs the target matrix. We believe that this does not occurs in an early stage of training, due to the prior information added by replacing the original patch-embedding layer with the pre-trained DINOv2 model. Despite of that, the use of earlier training checkpoints is essential, as an explicit regularization was not provided.



\section{Conclusion}
In this paper, we introduced a decrease-and-conquer strategy for the PlantCLEF 2025 challenge, where a Vision Transformer (ViT) was trained on a proxy task to perform segmentation previous to plant species identification. This strategy demonstrated competitive effectiveness, securing a 5th place on the private leaderboard of the challenge, with a final $F_1$-score of 0.33331. Our experiments confirmed that providing local context by assembling adjacent patches for classification is a crucial and highly effective strategy, significantly outperforming direct patch-wise classification. Given the narrow performance gap to top-ranking methods in the leaderboard, we are confident the proposed approach represents a promising direction for state-of-the-art multi-label classification in complex ecological imagery through domain-adaptation with further refinements holding the potential to achieve superior results.
\begin{acknowledgments}
  This work was partially supported by UEFS-AUXPPG 2023/2024/2025, CAPES-PROAP 2023/2024/2025, CAPES grant 88887.159255/2025-00 and  88887.594676/2020-00 and UEFS FINAPESQ (grant 047/2023).
\end{acknowledgments}

\section*{Declaration on Generative AI}
  
 During the preparation of this work, the author(s) used Chat-GPT-4o in order to: Grammar and spelling check. After using these tool(s)/service(s), the author(s) reviewed and edited the content as needed and take(s) full responsibility for the publication’s content.

\bibliography{sample-ceur}

@inproceedings{lifeclef2025,
title = {Overview of LifeCLEF 2025: Challenges on Species Presence Prediction and Identification, and Individual Animal Identification},
author = {Luk{\'a}{\v{s}} Picek and Stefan Kahl and Herv{\'e} Go{\"e}au and Luk{\'a}{\v{s}} Adam and Th{\'e}o Larcher and Cesar Leblanc and Maximilien Servajean and Kl{\'a}ra Janou{\v{s}}kov{\'a} and Ji{\v{r}}{\'i} Matas and Vojt{\v{e}}ch {\v{C}}erm{\'a}k and Kostas Papafitsoros and Robert Planqu{\'e} and Willem-Pier Vellinga and Holger Klinck and Tom Denton and Juan Sebasti{\'a}n Ca{\~n}as and Giulio Martellucci and Fabrice Vinatier and Pierre Bonnet and Alexis Joly},
booktitle = {International Conference of the Cross-Language Evaluation Forum for European Languages},
year = {2025},
organization = {Springer}
}

@inproceedings{plantclef2025,
author = {Giulio Martellucci and Herv{\'e} Go{\"e}au and Pierre Bonnet and Fabrice Vinatier and Alexis Joly},
title = {Overview of {PlantCLEF} 2025: Multi-Species Plant Identification in Vegetation Quadrat Images},
booktitle = {Working Notes of CLEF 2025 - Conference and Labs of the Evaluation Forum},
year = {2025}
}

@article{Bao2023transfer,
  title={A Survey of Heterogeneous Transfer Learning},
  author={Bao, Runze and Sun, Yuchen and Gao, Yifei and Wang, Jinhao and Yang, Qiao and Chen, Hong and Mao, Zhi-Hong and Ye, Yuyang},
  journal={arXiv preprint arXiv:2310.08459},
  year={2023},
  url={http://arxiv.org/abs/2310.08459},
  doi={10.48550/arXiv.2310.08459},
  note={arXiv:2310.08459 [cs]}
}

@inproceedings{Chulif2024neuon,
  author    = {Chulif, Sophia and Ishrat, Hamza Ahmed and Chang, Yang Loong and Lee, Sue Han},
  title     = {Patch-wise Inference using Pre-trained Vision Transformers: NEUON Submission to PlantCLEF 2024},
  booktitle = {Working Notes of CLEF 2024 - Conference and Labs of the Evaluation Forum},
  series    = {CEUR Workshop Proceedings},
  volume    = {3740}, 
  year      = {2024},
  note      = {CEUR-WS.org/Vol-3740/paper-192.pdf}
}

@inproceedings{Foy2024atlantic,
  author    = {Foy, Stephen and McLoughlin, Simon},
  title     = {Utilising DINOv2 for Domain Adaptation in Vegetation Plot Analysis},
  booktitle = {Working Notes of CLEF 2024 - Conference and Labs of the Evaluation Forum},
  series    = {CEUR Workshop Proceedings},
  volume    = {3740},
  year      = {2024},
  note      = {CEUR-WS.org/Vol-3740/paper-196.pdf}
}

@inproceedings{Goeau2024plantclef_overview,
  title={Overview of PlantCLEF 2024: Multi-species plant identification in vegetation plot images},
  author={Goëau, Hervé and Espitalier, Virgile and Bonnet, Pierre and Joly, Alexis},
  booktitle={Working Notes of CLEF 2024 - Conference and Labs of the Evaluation Forum},
  year={2024},
  organization={CEUR-WS.org}
}

@inproceedings{Joly2024lifeclef_overview,
  title={Overview of lifeclef 2024: Challenges on species distribution prediction and identification},
  author={Joly, Alexis and Picek, Lukáš and Kahl, Stefan and Goëau, Hervé and Espitalier, Virgile and Botella, Cédric and Deneu, Benjamin and Marcos, Diego and Estopinan, Janoma and Leblanc, Charles and Larcher, Titouan and Šulc, Milan and Hrúz, Matúš and Servajean, Maximilien and Matas, Jiri and others},
  booktitle={International Conference of the Cross-Language Evaluation Forum for European Languages},
  year={2024},
  organization={Springer}
}

@article{Oquab2023dinov2,
  title={DINOv2: Learning Robust Visual Features without Supervision},
  author={Oquab, Maxime and Darcet, Timothée and Moutakanni, Théo and Vo, Huy V and Szafraniec, Marc and Khalidov, Vasil and Fernandez, Pierre and Haziza, Daniel and Massa, Francisco and El-Nouby, Alaaeldin and Assran, Mahmoud and Ballas, Nicolas and Galuba, Wojciech and Howes, Russell and Huang, Po-Yao and Li, Shang-Wen and Misra, Ishan and Rabbat, Michael and Sharma, Vasu and Synnaeve, Gabriel and Xu, Hu and Jégou, Hervé and Mairal, Julien and Labatut, Patrick and Joulin, Armand and Bojanowski, Piotr},
  journal={arXiv preprint arXiv:2304.07193},
  year={2023}
}

@misc{Goeau2024models,
  author={Goëau, Hervé and Lombardo, Jean-Christophe and Affouard, Antoine and Espitalier, Virgile and Bonnet, Pierre and Joly, Alexis},
  title={PlantCLEF 2024 pretrained models on the flora of the south western Europe based on a subset of Pl@ntNet collaborative images and a ViT base patch 14 dinoV2},
  year={2024},
  publisher={Zenodo},
  doi={10.5281/zenodo.10848263},
  howpublished={\url{https://doi.org/10.5281/zenodo.10848263}}
}

@article{Kattenborn2021cnn,
  title={Review on convolutional neural networks (cnn) in vegetation remote sensing},
  author={Kattenborn, Teja and Leitloff, Jana and Schiefer, Franziska and Hinz, Stefan},
  journal={ISPRS journal of photogrammetry and remote sensing},
  volume={173},
  pages={24--49},
  year={2021},
  publisher={Elsevier}
}

@inproceedings{Caron2021emerging,
  title={Emerging Properties in Self-Supervised Vision Transformers},
  author={Caron, Mathilde and Touvron, Hugo and Misra, Ishan and Jégou, Hervé and Mairal, Julien and Bojanowski, Piotr and Joulin, Armand},
  booktitle={Proceedings of the IEEE/CVF International Conference on Computer Vision (ICCV)},
  pages={9650--9660},
  year={2021}
}

@inproceedings{ijepa,
  title={Self-supervised learning from images with a joint-embedding predictive architecture},
  author={Assran, Mahmoud and Duval, Quentin and Misra, Ishan and Bojanowski, Piotr and Vincent, Pascal and Rabbat, Michael and LeCun, Yann and Ballas, Nicolas},
  booktitle={Proceedings of the IEEE/CVF Conference on Computer Vision and Pattern Recognition},
  pages={15619--15629},
  year={2023}
}

@String{Computer = "{IEEE} Computer" }

@String{Springer = "Springer-Verlag" }

\appendix



\end{document}